\documentclass[conference]{IEEEtran}
\usepackage[style=ieee, backend=biber]{biblatex}
\usepackage{amsmath,amssymb,amsfonts}
\usepackage{algorithmic}
\usepackage{graphicx}
\usepackage{textcomp}
\usepackage{xcolor}
\usepackage{subcaption}
\usepackage{pgfplots}
\pgfplotsset{compat=1.18}
\usepackage{pgfplotstable}
\usepackage{booktabs}
\usepackage{float}
\usepackage{cuted}
\usepackage{eso-pic}

\AddToShipoutPictureFG*{%
  \AtTextLowerLeft{%
    \raisebox{-1.5cm}{%
      \parbox{\textwidth}{%
        \centering
        \fontsize{7}{8}\selectfont
        © 2025 IEEE. Personal use of this material is permitted. Permission from IEEE must be obtained for all other uses, in any current or future media, including reprinting/republishing this material for advertising or promotional purposes, creating new collective works, for resale or redistribution to servers or lists, or reuse of any copyrighted component of this work in other works. This is the Accepted Author Manuscript of the paper published in the Proceedings of the International Joint Conference on Neural Networks (IJCNN 2025). The version of record is available at DOI: 10.1109/IJCNN64981.2025.11227194.
      }%
    }%
  }%
}

\def\BibTeX{{\rm B\kern-.05em{\sc i\kern-.025em b}\kern-.08em
    T\kern-.1667em\lower.7ex\hbox{E}\kern-.125emX}}

\begin{document}

\title{LLM-Based Embeddings for Program Analysis and Optimization}

\author{\IEEEauthorblockN{Anonymous Authors}}

\author{\IEEEauthorblockN{Calvin Higgins}
\IEEEauthorblockA{\textit{Department of Computer Science and Statistics} \\
\textit{University of Rhode Island}\\
Kingston, United States of America \\
calvin\_higgins2@uri.edu}
\and
\IEEEauthorblockN{Marco Alvarez}
\IEEEauthorblockA{\textit{Department of Computer Science and Statistics} \\
\textit{University of Rhode Island}\\
Kingston, United States of America \\
malvarez@cs.uri.edu}
}

\maketitle

\begin{abstract}
Recent advances have highlighted the potential of machine learning, particularly Large 
Language Models (LLMs), for analyzing and optimizing programs. We present the first 
application of program embeddings from LLMCompiler---an LLM massively pretrained on 
intermediate representation (IR) code---to representative program analysis and 
optimization tasks. We generate program embeddings directly from source and IR code 
using a simple approach: split programs into chunks, independently embed each chunk with
pretrained LLMs, and then aggregate the chunk embeddings into a single program 
embedding. Our experiments show that combining source and IR code embeddings achieves an
error rate of 1.54\% in algorithm classification, a 12\% improvement over the current 
state-of-the-art, and a competitive accuracy on heterogeneous device mapping. These 
findings suggest that training a performance-aware LLM for embedding IR code might yield
state-of-the-art results in code optimization tasks.
\end{abstract}


\section{Introduction}

Software developers are producing code at an unprecedented rate. In 2024, GitHub saw 
5.2 billion new contributions and reported widespread adoption of 
generative AI coding tools\footnote{https://octoverse.github.com}. Scaling both program
analysis (e.g. algorithm classification) and optimization tasks (e.g. heterogeneous 
device mapping) accordingly requires automated tooling.

One current machine learning approach constructs semantic program embeddings and applies 
them to downstream program analysis and optimization tasks. Instead of embedding source 
or machine code directly, many methods remain language and architecture-agnostic by 
embedding intermediate representation (IR) code. This IR can be viewed as a virtual 
assembly language: a standardized, low-level abstraction of program behavior that is 
independent of source language and machine architecture~\cite{chris-lattner-2004}. 
An alternative approach, rather than operating on textual IR directly, constructs graph 
representations from IR code and then applies Graph Neural Networks 
(GNNs)~\cite{akash-dutta-2023, changan-niu-2024, jamsaz-ali-2024}.

Although there has been some work on text-based IR models, such as 
OSCAR~\cite{dinglan-peng-2021} graph-based and multimodal methods have largely 
outperformed textual models~\cite{akash-dutta-2023, changan-niu-2024, jamsaz-ali-2024}. 
Graph-structured inductive biases have proven highly effective for learning program 
embeddings. However, until recently, no large-scale text-based IR model had employed 
massive pretraining like their source code counterparts. In~\cite{chris-cummins-2025}, 
Cummins et al. introduced LLMCompiler, a 70-billion parameter Large Language Model (LLM)
pretrained on 546 billion tokens of LLVM IR and assembly code. LLVM IR is an IR defined 
by the LLVM compiler infrastructure~\cite{chris-lattner-2004}. To date, program 
embeddings from LLMCompiler have not been applied to standard downstream program 
analysis and optimization tasks. 

Inspired by LLMCompiler, we employ pretrained LLMs to generate program embeddings from
source and IR code. These two views complement each other: source code captures 
high-level algorithmic intent, while IR code exposes low-level data and control flow. To 
embed programs, we first divide them into chunks, embed each chunk independently, and 
then aggregate them into a program embedding with a Long-Short Term Memory (LSTM) 
network. 

The main contributions of this work are:
\begin{enumerate}
    \item A lightweight embedding framework that feeds pretrained LLMs chunked source
    and IR code, while fine-tuning only a LSTM head.
    \item The first empirical evaluation of LLMCompiler-based embeddings on 
    representative downstream tasks in program analysis and optimization.
    \item State-of-the-art performance on algorithm classification and competitive 
    accuracy on heterogeneous device mapping.
\end{enumerate}

These results indicate that massive pretraining can close the performance gap between
graph-based and text-based models. Performance-aware pretraining of LLMCompiler and
improved embedding methods remain an open path to further gains on program analysis and 
optimization tasks. 

The remainder of this paper is organized as follows. Section~\ref{section:related_work} 
provides an overview of related work in machine learning for program analysis and 
optimization, followed by a description of our methodology in 
Section~\ref{section:methods}. Sections~\ref{section:hdm} and \ref{section:ac} present 
our experimental setup, results and discuss key findings, and finally, 
Section~\ref{section:conclusion} summarizes contributions and potential future 
directions.
\section{Related Work}
\label{section:related_work}

Graph-based representations are central to modern program analysis and optimization,
particularly Control Flow Graphs (CFGs) and Data Flow Graphs (DFGs). CFGs model a 
program's execution paths: each node is a basic block of instructions, and directed 
edges indicate possible control flow between blocks. DFGs capture how values propagate 
through a program: nodes denote operations or value definitions, and directed edges 
connect every producer to the operations that consume its results.    

Although plain CFGs and DFGs are effective, they omit crucial semantics such as operand 
order, memory aliasing and interprocedural context, motivating a rich lineage of 
enhanced graph representations. ConteXual Flow Graphs (XFGs), used by inst2vec, merge 
CFGs and DFGs into a single graph~\cite{tal-ben-nun-2018}. Control Data Flow Graphs 
(CDFGs) extend XFGs with explicit memory dependency edges between load and store 
instructions~\cite{alexander-brauckmann-2020}. ProGraML unifies control, data and call 
flow while encoding operand positions and literal values~\cite{chris-cummins-2021}. 
PerfoGraph extends ProGraML with aggregate data type information~\cite{jamsaz-ali-2024}.
Finally, FAIR constructs separate CFGs and DFGs augmented with call 
flow information~\cite{changan-niu-2024}.

Although these enhanced graphs improve downstream performance, they require careful 
feature engineering. To avoid this, some methods operate directly on 
textual IR to automatically extract relevant features but without the inductive biases
of graph representations they generally lag behind graph-based models. OSCAR, a 
hierarchical Transformer trained with textual IR, delivers competitive results but still
falls short of the best graph-based models~\cite{dinglan-peng-2021}. MIREncoder adopts
a multimodal approach, incorporating textual IR as well as data, control and call flow 
graphs, achieving strong performance~\cite{akash-dutta-2024}. Both models, however, are 
orders of magnitude smaller than today's large-scale source-code LLMs, which limits 
their capacity to generalize.

LLMCompiler narrows the scale gap with a 70-billion parameter LLM pretrained on 546B 
LLVM IR and assembly tokens~\cite{chris-cummins-2025}. Yet its embeddings have never 
been evaluated on representative program analysis or optimization tasks, so we still do 
not know whether massive pretraining on textual IR can rival graph-based embedding 
methods. This paper fills that gap: we extract program embeddings from LLMCompiler and 
deliver the first evaluation on algorithm classification and heterogeneous device 
mapping.
\section{Methods and Approach}
\label{section:methods}

This work presents a comparative analysis of program embeddings from pretrained LLMs 
and GNNs. The evaluation covers two classification tasks: heterogeneous device mapping 
and algorithm classification. This section describes the embedding generation methods.

\begin{figure*}
    \centering
    \includegraphics[width=.85\linewidth]{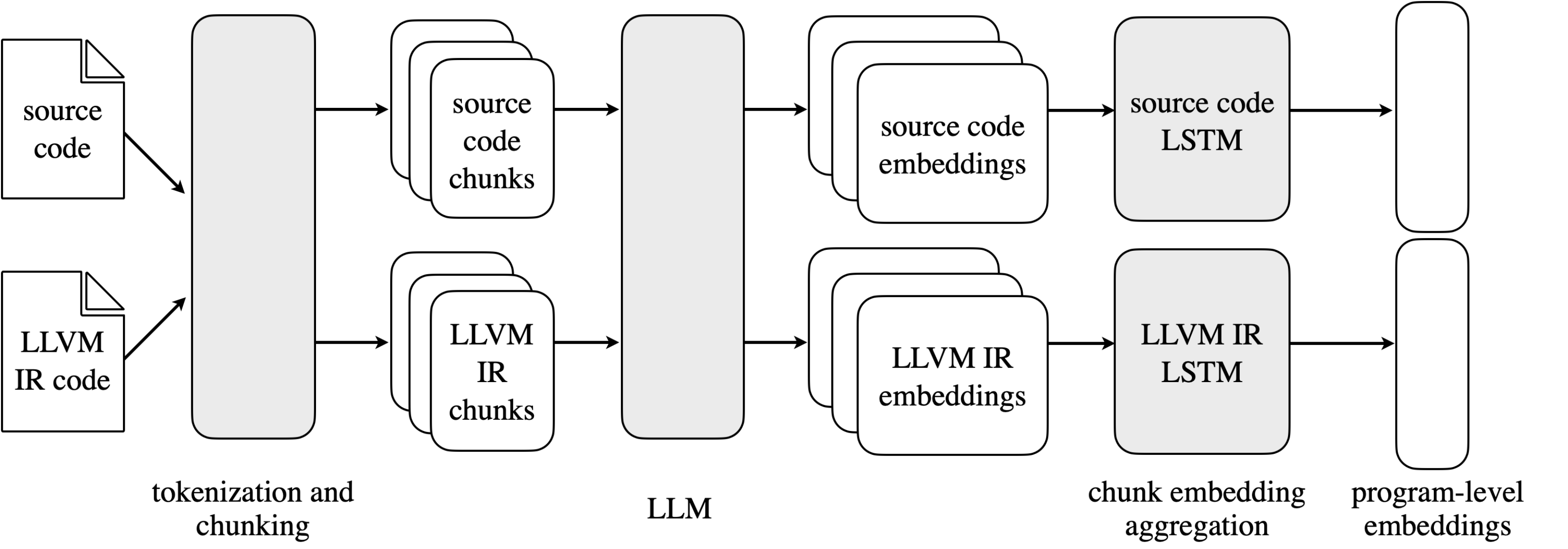}
    \caption{The program embedding procedure. We use the same pretrained LLM to 
    generate chunk embeddings for the source and LLVM IR code. The chunk embeddings are 
    aggregated by separate LSTMs and concatenated to produce a program-level embedding. 
    Note that while the LSTMs are trainable, the LLM is not fine-tuned.}
    \label{fig:model}
\end{figure*}

\subsection{Program Graphs}

For GNN approach, programs are modeled as control flow graphs (CFGs). A CFG is a 
directed graph $G = (V, E)$, where $V$ is the set of basic blocks (nodes) and 
$E \subseteq V \times V$ denotes control flow transitions between basic blocks (edges).
Every node in $V$ can be labeled with the following feature vectors:

\paragraph{Instruction counts}
A histogram of the number of instructions in the basic block.
\paragraph{Autophase features}
A vector with feature counts for the following 
autophase features~\cite{ameer-haj-ali-2020-id62}: 
\textit{ArgsPhi, BBPhi, BeginPhi, BinaryConstArg, Const32Bit, Const64Bit, NumConstOnes,
NumConstZeroes, TestUnary, TotalInsts, TotalMemInst}. 

\subsection{Graph Neural Networks}

We experiment with two GNN architectures that have shown state-of-the-art performance on
various tasks:

\paragraph{Graph Attention Network} 
The GAT~\cite{shaked-brody-2022} formulation incorporates learned attention mechanisms 
over node neighborhoods through the following message passing scheme:
\begin{displaymath}
    h'_i = \sigma\left(\sum_{j \in \mathcal{N}_i} \alpha_{ij} W h_j\right)
\end{displaymath}
where $\alpha_{ij}$ are learned attention coefficients, $W$ are network parameters, and 
$\mathcal{N}_i$ is the neighborhood of node $i$. 

\paragraph{Principal Neighborhood Aggregation}
PNA~\cite{gabriele-corso-2020} extends traditional GNNs with continuous features and 
degree-aware aggregators, enabling the network to adjust signals based on the degree of 
each node. Empirical results support PNA's effectiveness in capturing graph structures 
and outperforming other GNN architectures.

\subsection{Pretrained LLMs}

Transfer learning leverages models pretrained on large datasets to extract features for 
downstream tasks. We use the following pretrained LLMs to generate program embeddings.

\paragraph{CodeBERT~\cite{zhangyin-feng-2020}}
CodeBERT is an encoder-only LLM that uses masked language modeling and replaced token 
detection to capture syntactic and semantic nuances. Its dual pretraining on natural 
and programming languages enhances contextual understanding and performance in tasks 
like code summarization and search.

\paragraph{CodeLlama~\cite{baptiste-roziere-2024}}
Building on Llama2, CodeLlama is a decoder-only LLM pretrained and fine-tuned for tasks
like code synthesis and analysis. 

\paragraph{LLMCompiler~\cite{chris-cummins-2025}}
Building on CodeLlama, LLMCompiler is a decoder-only LLM designed for code 
optimization. LLMCompiler's pretraining and fine-tuning on 546B tokens of LLVM IR and 
assembly code, improves its understanding of compiler IRs, assembly language, and 
optimization techniques. 

\paragraph{Llama3}
Llama3\footnote{https://ai.meta.com/blog/meta-llama-3/} is a standard decoder-only LLM 
with an efficient tokenizer, leading to improved model performance. Grouped query 
attention is used to reduce inference cost.

\subsection{Generating Code Embeddings from LLMs}

Given the challenges of processing large programs with LLMs, we adopt a chunk-based 
approach, illustrated in Fig.~\ref{fig:model} to generate the embeddings. First we 
tokenize the input program using the LLM's default tokenizer. As programs may exceed the
LLM's context window (the maximum number of tokens the LLM can process at once), we 
divide input programs into fixed-size, non-overlapping chunks. Although overlapping 
chunks offer additional context, we opt for non-overlapping chunks to reduce the number 
of forward passes through the LLM and improve computational efficiency. 

Each chunk of tokens is processed independently by the LLM to generate token-level 
embeddings. The embedding for each token is extracted from the layer just before the LLM
head. To obtain a chunk-level embedding, we either use the last token embedding in the 
chunk or compute the mean of all token embeddings in the chunk. These chunk-level 
embeddings are then aggregated into a single program-level embedding using a 
long-short-term memory (LSTM) network. We choose an LSTM over simple aggregation 
strategies, such as mean aggregation, to capture dependencies between chunks, 
particularly when important relationships in the code may span multiple chunks. 

For our experiments, we compute program embeddings from both the source and LLVM IR 
code. The source and LLVM IR code are processed independently by two separate LSTMs. 
This provides an embedding of both the high-level and low-level program representations.
Having both embeddings offers several advantages: the high-level source code captures 
the program's original intent, structure, and logic, making it easier to understand the 
program's functionality, while the low-level LLVM IR offers a more detailed view of how 
the code will be executed on the hardware, including optimizations and instruction-level
nuances. Combining these two perspectives ensures that the model benefits from both 
abstract and concrete representations of the program. For downstream classification 
tasks, both program-level embeddings are concatenated with any additional global program
features and passed to a classifier head, a multi-layer perceptron (MLP), to produce the
final classification.
\section{Heterogeneous Device Mapping}
\label{section:hdm}
This section evaluates the proposed models on heterogeneous device mapping, a 
representative program optimization task, comparing GNNs, LLMs, and state-of-the-art 
approaches.

\subsection{Dataset}

Heterogeneous device mapping involves classifying whether an OpenCL kernel runs faster
on a CPU or a GPU. The dataset contains 680 labeled examples derived from 256 distinct 
kernels across seven benchmark suites (\textit{AMD SDK, NPB, NVIDIA SDK, Parboil, 
Polybench, Rodinia, SHOC})~\cite{dominik-grewe-2013}. Some kernels appear multiple times
with varied runtime parameters, including data transfer size, workgroup size (threads 
per workgroup), and chosen dataset (e.g. sorted or unsorted), as optimal device 
placement depends on both the code and runtime settings. Labels are provided for AMD 
GPU (Tahiti 7970) vs. Intel CPU (i7-3820) and NVIDIA GPU (GTX 970) vs. Intel CPU 
(i7-3820) performance. 

\subsection{GNN Preprocessing}
\label{section:hdm_gnn_pre}

CFGs are built from the LLVM IR code with instruction count and autophase node features. 
Global features include transfer size, workgroup size, their log-transformed values
($\log(x+1)$) and a one-hot encoding of the chosen dataset. Global features with zero 
variance are removed, and the remaining features are standardized to have zero mean
and unit variance.

\subsection{LLM Preprocessing}
\label{section:hdm_llm_pre}

The OpenCL source code is preprocessed as described in \cite{chris-cummins-2017}, while 
only the instructions and labels from the LLVM IR code are retained. Both are tokenized 
and chunked. For decoder-only LLMs, the first chunk begins with a beginning-of-sequence 
token (BOS) and the last chunk ends with an end-of-sequence token (EOS). For 
encoder-only LLMs, each chunk begins with a classification token (CLS) and ends with an 
EOS token. The last chunk is always padded with EOS tokens which are ignored during 
token embedding aggregation. Global features are preprocessed as described in 
Section~\ref{section:hdm_gnn_pre} and passed to the classifier head.

\subsection{Stratified Nested Cross Validation}
\label{section:hdm_str_ncv}

Instead of the standard 10-fold stratified cross-validation (CV), we use 5-fold outer 
and 10-fold inner stratified nested cross-validation (stratified NCV) to reduce 
variance. In the inner sweep, models are trained with early stopping (patience 10 
epochs, minimum delta 0.0001) for up to 250 epochs. The inner hyperparameter combination
with maximum validation Matthews correlation coefficient (MCC) is retrained for a fixed 
number of epochs on the combined training and validation sets. The number of training 
epochs is scaled according to the ratio of the combined dataset size (training + 
validation) to the original training set size. The retrained model is evaluated on the 
test set, and its predictions are saved. Test metrics for outer hyperparameter 
combinations are calculated by aggregating these predictions from each inner fold, with 
the best outer hyperparameter combination selected according to maximum test accuracy.

Models are trained with the AdamW optimizer ($\beta_1 = 0.9$, $\beta_2 = 0.999$,
$\lambda = 0.01$), a batch size of 256, and cosine learning rate warmup for 10 steps. 
The classifier head is a multi-layer perceptron (MLP) with two hidden layers: the first 
is twice the input size and the second matches the input size. Batch normalization and 
ReLU activation are applied between layers, except for the last. Other hyperparameters 
are selected via hyperparameter sweeps.

\paragraph{GNN Hyperparameters}
\label{section:hdm_gnn_hps}

We evaluate the GAT and PNA architectures. The outer sweep tunes data transfer and 
workgroup size features (enabled/disabled), chosen dataset feature (enabled/disabled), 
instruction counts (enabled/disabled), autophase features (enabled/disabled), and GNN 
aggregator (Max/Min/MLP). The inner sweep tunes learning rate (0.005/0.0005), dropout 
rate (0.0/0.25), GNN hidden size (32/64), and GNN layers (1/2/4).

\paragraph{LLM Hyperparameters}
\label{section:hdm_llm_hps}

We evaluate six decoder-only models (CodeLlama-7B, CodeLlama-70B, Llama3-8B, Llama3-70b,
LLM-Compiler-7B, LLM-Compiler-13B) and one encoder-only model (CodeBERT-125M). For 
decoder-only models, the outer sweep tunes chunk size (4096/16384) and aggregator 
(Last/Mean). For encoder-only models, it tunes chunk size (512) and aggregator 
(First/Last/Mean). The inner sweep tunes learning rate (0.005/0.0005), dropout rate 
(0.5), LSTM hidden size (32/64), LSTM layers (1/2), and LSTM bidirectional 
(enabled/disabled).

\paragraph{Results}

Examining Table~\ref{tab:ncv_hdm}, we observe that the decoder-only 
LLMs outperform the GNN baselines for both the AMD and NVIDIA labels. Surprisingly, 
LLMCompiler, the only model pretrained on both source and LLVM IR code, achieves similar
accuracy to the generic Llama variants. 

A likely reason lies in the mismatch between LLMCompiler's pretraining tasks and the 
demands of heterogeneous device mapping. LLMCompiler is initialized from CodeLlama 
then finetuned with autoregressive language modeling on a large corpus of LLVM IR
and assembly. LLMCompiler is also trained to emulate LLVM passes: given unoptimized 
program, starting code size and an optimization flag list, output the optimized program 
and code size. The sole explicit performance cue is code size. Critical 
hardware-dependent factors such as latency, bandwidth and instruction throughput never 
appear in the loss. As a result, LLMCompiler learns syntactic and optimization patterns 
but might not fully capture code runtime in its embeddings. 

\begin{table}[htbp]
    \centering
    \caption{Performance with stratified NCV. Numbers in parentheses are the weighted 
    standard deviation of the per-fold accuracies.}
    \label{tab:ncv_hdm}
    \begin{tabular}{ccc}
        \toprule
        Model               & AMD Acc. & NVIDIA Acc. \\
        \midrule
        PNA                 & 88.38 (2.48) & 87.65 (1.50) \\
        GAT                 & 88.68 (1.51) & 86.32 (0.59) \\
        \midrule
        CodeLlama-7B        & \textbf{92.06 (1.99)} & \textbf{88.38 (2.44)} \\
        CodeLlama-70B       & 91.03 (2.39) & 85.74 (2.40) \\
        Llama3-8B           & 91.18 (2.28) & 86.76 (2.67) \\
        Llama3-70B          & 91.32 (1.94) & 85.00 (2.31) \\
        LLMCompiler-7B      & 91.18 (2.98) & 87.50 (1.32) \\
        LLMCompiler-13B     & 91.03 (1.88) & 86.62 (1.89) \\
        CodeBERT            & 87.50 (3.05) & 82.21 (4.09) \\
        \bottomrule
    \end{tabular}
\end{table}

\subsection{Leave-One-Group-Out Nested Cross Validation}
\label{section:logo_ncv}

Stratified CV presents challenges for this dataset, as identical source code may appear
in both the training and test sets because kernels are repeated with varied runtime 
parameters. Additionally, kernels from the same benchmark suite, which might be similar 
due to co-evolution, can also appear in both sets. This risks test set leakage,
potentially inflating results from prior literature using stratified CV.

To illustrate this, we reinterpret our 5-fold outer, 10-fold inner stratified NCV 
results as five different 10-fold stratified CV runs. We take the validation accuracy of
the best model as the ``test'' accuracy. Histograms of the test accuracies for AMD and 
NVIDIA labels are shown in Fig.~\ref{fig:stratified_cv_histogram_amd} and 
Fig.~\ref{fig:stratified_cv_histogram_nvidia}. 

\begin{figure}[htbp]
    \centering
    \begin{tikzpicture}[scale=.85]
        \begin{axis}[
            ybar,
            xlabel={AMD Test Accuracy},
            ylabel={Frequency},
            ymin=0,
            grid=both,
            major grid style={line width=0.2pt,draw=gray!50},
        ]
        \addplot+[ybar interval,mark=no] plot coordinates { (66.54, 2) (67.68, 3) (68.82, 12) (69.96, 14) (71.1, 18) (72.24, 21) (73.38, 24) (74.52, 41) (75.66, 55) (76.8, 52) (77.94, 67) (79.08, 38) (80.22, 48) (81.36, 25) (82.5, 43) (83.64, 40) (84.78, 38) (85.92, 59) (87.06, 66) (88.2, 62) (89.34, 127) (90.48, 217) (91.62, 311) (92.76, 256) (93.9, 16) };
        \end{axis}
    \end{tikzpicture}
    \caption{Accuracies across five different 10-fold stratified CV splits with the AMD labels.}
    \label{fig:stratified_cv_histogram_amd}
\end{figure}
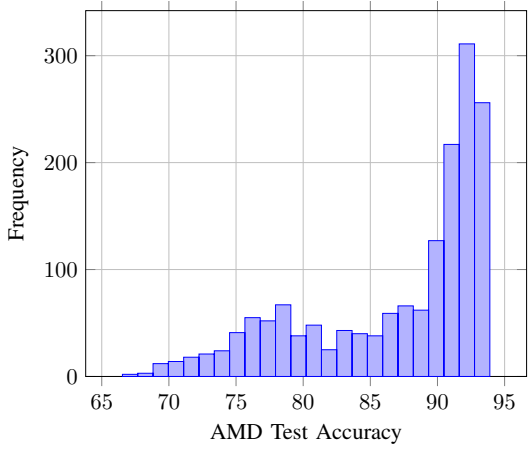
\begin{figure}[htbp]
    \centering
    \begin{tikzpicture}[scale=.85]
        \begin{axis}[
            ybar,
            xlabel={NVIDIA Test Accuracy},
            ylabel={Frequency},
            ymin=0,
            grid=both,
            major grid style={line width=0.2pt,draw=gray!50},
        ]
        \addplot+[ybar interval,mark=no] plot coordinates { (61.4, 2) (62.64, 1) (63.87, 1) (65.11, 3) (66.34, 5) (67.58, 1) (68.81, 5) (70.05, 21) (71.28, 24) (72.52, 29) (73.75, 42) (74.99, 23) (76.22, 33) (77.46, 44) (78.69, 59) (79.93, 95) (81.16, 123) (82.4, 97) (83.63, 103) (84.87, 149) (86.1, 224) (87.34, 245) (88.57, 237) (89.81, 85) (91.04, 9) };
        \end{axis}
    \end{tikzpicture}
    \caption{Accuracies across five different 10-fold stratified CV splits with the NVIDIA labels.}
    \label{fig:stratified_cv_histogram_nvidia}
\end{figure}
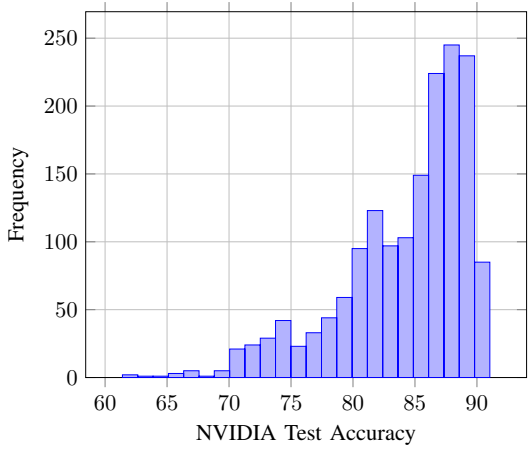

Notably, when compared to the results from stratified NCV, the maximum AMD test accuracy
increased from 92.06\% to 95.04\%, and the maximum NVIDIA test accuracy rose from 
88.38\% to 92.28\%. By considering five random splits, our 10-fold stratified CV results
meet the state-of-the-art, underscoring the impact of test-set leakage and the need for 
grouped splits. 

To mitigate these issues, we use leave-one-group-out nested cross-validation (LOGO NCV). 
With seven groups, this results in seven outer folds and six inner folds. As the 
group sizes are highly imbalanced, we also report the weighted standard deviation of the
per-fold accuracies. The training and model selection procedures are otherwise 
consistent with Section \ref{section:hdm_str_ncv}.

\paragraph{Results}

In Table~\ref{tab:logo_ncv_hdm}, we observe that the LLMs are competitive with the GNN
baselines on AMD labels, although the GNNs outperform the LLMs on NVIDIA labels. While
this could be due to the high variance caused by imbalanced folds, its likely caused by
overfitting, as discussed in Section~\ref{section:hdm_ablation}.

\begin{table}[htbp]
    \centering
    \caption{Performance with LOGO NCV. Numbers in parentheses are the weighted standard
    deviation of the per-fold accuracies. }
    \label{tab:logo_ncv_hdm}
    \begin{tabular}{ccc}
        \toprule
        Model & AMD Acc. & NVIDIA Acc. \\
        \midrule
        PNA                 & 72.65 (5.87) & \textbf{60.59 (17.52)} \\
        GAT                 & 75.44 (4.68) & 56.76 (8.80) \\
        \midrule
        CodeLlama-7B        & 75.29 (7.79) & 46.62 (6.94) \\
        CodeLlama-70B       & 71.47 (7.00) & 44.12 (5.01) \\
        Llama3-8B           & 74.85 (5.14) & 50.88 (5.57) \\
        Llama3-70B          & 74.12 (4.82) & 45.88 (8.33) \\
        LLMCompiler-7B      & \textbf{76.03 (5.15)} & 46.76 (12.53) \\
        LLMCompiler-13B     & 72.65 (7.65) & 50.74 (14.66) \\
        CodeBERT            & 73.24 (8.68) & 43.82 (10.39) \\
        \bottomrule
    \end{tabular}
\end{table}

\subsection{State-of-the-Art Approaches}

To compare our approach with the state-of-the-art, we select the LLM and GNN models that
achieved the highest test accuracy in stratified NCV. As shown in 
Table~\ref{tab:ncv_hdm_others}, while our LLM method is competitive, it doesn't achieve 
the state-of-the-art. However, it's important to note that the other methods report 
results using 10-fold CV or 10-fold stratified CV whereas we employ a more robust 5-fold 
outer, 10-fold inner stratified NCV. This approach reduces the risk of performance 
being skewed by a favorable data split. 

\begin{table}[htbp]
    \centering
    \caption{Comparison with other methods.}
    \label{tab:ncv_hdm_others}
    \begin{tabular}{cccc}
        \toprule
        Model & AMD Acc. & NVIDIA Acc. & CV Method \\
        \midrule
        GNN (ours)                                      & 88.68 & 87.65 & Stratified NCV \\
        LLM (ours)                                      & 92.06 & 88.38 & Stratified NCV \\
        LLM (ours)                                      & 95.04 & 92.28 & Stratified CV  \\
        \midrule
        inst2vec-imm~\cite{tal-ben-nun-2018}            & 88.09 & 86.62 & Stratified CV \\
        OSCAR~\cite{dinglan-peng-2021}                  & 88.8 & 89.7 & Stratified CV \\
        ProGraML~\cite{chris-cummins-2021}              & 86.6 & 80.0 & CV \\
        PerfoGraph~\cite{jamsaz-ali-2024}               & 94 & 90 & CV \\
        FAIR~\cite{changan-niu-2024}                    & \textbf{96.52} & 91.61 & CV \\
        MIREncoder~\cite{akash-dutta-2024}              & 93.6 & \textbf{93.7} & Stratified CV \\
        \bottomrule
    \end{tabular}
\end{table}

Additionally, the potential for test-set leakage is significant without a grouped split, 
raising concerns about the reliability of stratified CV results. In 
Section~\ref{section:logo_ncv}, we demonstrate this effect and show that our method 
meets the state-of-the-art simply by trying five different 10-fold stratified CV 
splits. Moreover, the high variance of stratified split results is already reflected in 
the existing literature. For example, the authors of~\cite{changan-niu-2024} report that
ProGraML achieves 92.60\% AMD accuracy and 88.13\% NVIDIA accuracy on their split while 
the original ProGraML paper reports only 86.6\% AMD accuracy and 80.0\% NVIDIA accuracy 
on a different split~\cite{chris-cummins-2021}. 

Unfortunately only \cite{alexander-brauckmann-2020} and \cite{emanuele-parisi-2022} have
considered a grouped split. The authors of \cite{alexander-brauckmann-2020} also 
evaluate several existing but older methods shown in 
Table~\ref{tab:logo_ncv_hdm_others}. Of the results available in the literature, ours 
are the best, although we suspect modern methods would achieve similar results.

\begin{table}[htbp]
    \centering
    \caption{Comparison with other methods. Results marked with $*$ are 
    from~\cite{alexander-brauckmann-2020} and not the original paper.}
    \label{tab:logo_ncv_hdm_others}
    \begin{tabular}{cccc}
        \toprule
        Model & AMD Acc. & NVIDIA Acc. & CV Method \\
        \midrule
        GNN (ours)                                      & 75.44 & \textbf{60.59} & LOGO NCV \\
        LLM (ours)                                      & \textbf{76.03} & 50.88 & LOGO NCV \\
        \midrule
        Grewe et al.~\cite{dominik-grewe-2013}          & $56^*$ & $38^*$ & LOGO CV \\ 
        DeepTune~\cite{chris-cummins-2017}              & $48^*$ & $48^*$ & LOGO CV \\
        inst2vec~\cite{tal-ben-nun-2018}                & $49^*$ & $41^*$ & LOGO CV \\
        GNN-CDFG~\cite{alexander-brauckmann-2020}       & 51 & 46 & LOGO CV \\
        GNN-AST~\cite{alexander-brauckmann-2020}        & 64 & 58 & LOGO CV \\
        CNN~\cite{emanuele-parisi-2022}                 & 49.2 & 44.3 & LOGO CV \\
        \bottomrule
    \end{tabular}
\end{table}

Given this, it's uncertain which method truly outperforms the others, as the potential 
for bias in the reported results makes direct comparison difficult. However, the results 
in Table~\ref{tab:ncv_hdm_others} and Table~\ref{tab:logo_ncv_hdm_others} show that 
our method is at least competitive with the state-of-the-art.

\begin{table*}
    \centering
    \caption{Input features ablation with LOGO NCV}
    \label{tab:hdm_feat_abl}
    \begin{tabular}{ccccccccc}
        \toprule
        & \multicolumn{4}{c}{No Source Code} & \multicolumn{4}{c}{No LLVM IR Code} \\
        \cmidrule(r){2-5}
        \cmidrule(r){6-9}
        
        & \multicolumn{2}{c}{AMD} & \multicolumn{2}{c}{NVIDIA} & \multicolumn{2}{c}{AMD} & \multicolumn{2}{c}{NVIDIA} \\
        \cmidrule(r){2-3}
        \cmidrule(r){4-5}
        \cmidrule(r){6-7}
        \cmidrule(r){8-9}
        
        Model & Acc. & Change & Acc. & Change & Acc. & Change & Acc. & Change \\
        \midrule
        CodeLlama-7B        & 71.32 & -5.27 & 43.97 & -5.68 & 72.79 & -3.32 & 42.06 & -9.78 \\
        CodeLlama-70B       & 70.74 & -1.02 & 40.74 & -7.66 & 72.94 & \textbf{2.06} & 40.29 & -8.68 \\
        Llama3-8B           & 71.62 & -4.32 & 40.59 & -20.22 & 72.21 & -3.53 & 46.76 & -8.10 \\
        Llama3-70B          & 70.59 & -4.76 & \textbf{44.41} & \textbf{-3.20} & 73.53 & -0.80 & 50.44 & 9.94 \\
        LLMCompiler-7B      & 68.97 & -9.29 & 42.21 & -9.73 & \textbf{74.56} & -1.93 & \textbf{54.12} & \textbf{15.74} \\
        LLMCompiler-13B     & 69.26 & -4.67 & 41.91 & -17.38 & 70.29 & -3.25 & 42.94 & -15.37 \\
        CodeBERT            & \textbf{72.79} & \textbf{-0.61} & 41.32 & -5.71 & 72.94 & -0.41 & 43.24 & -1.32 \\
        \bottomrule
    \end{tabular}
\end{table*}

\subsection{Ablation}
\label{section:hdm_ablation}

After removing the source code, accuracy decreases across all models, indicating that it
contains relevant information. Likewise, after removing the LLVM IR code, accuracy 
decreases for most models, although a few become slightly more accurate. In general, the
results in Table~\ref{tab:hdm_feat_abl}, indicate that including both the source and 
LLVM IR code improves performance. That is, the high-level intent captured by the source
code and the low level details captured by the LLVM IR code are both relevant to 
heterogeneous device mapping.

We also tried replacing the LSTMs with unweighted mean aggregation. As shown in 
Table~\ref{tab:hdm_lstm_abl}, accuracy decreased for AMD labels but increased for NVIDIA
labels. Notably, the LLMs achieve NVIDIA accuracy comparable to PNA, the best GNN model. 
This suggests that the LSTMs might be overfitting the NVIDIA labels due to limited 
training data in some imbalanced folds. 

\begin{table}[htbp]
    \centering
    \caption{LSTM ablation with LOGO NCV}
    \label{tab:hdm_lstm_abl}
    \begin{tabular}{ccccc}
        \toprule
        & \multicolumn{2}{c}{AMD} & \multicolumn{2}{c}{NVIDIA} \\
        Model & Acc. & Change & Acc. & Change \\
        \midrule
        CodeLlama-7B        & 66.18 & -12.30 & 59.26 & 27.11 \\
        CodeLlama-70B       & 63.24 & -11.52 & 60.15 & 36.33 \\
        Llama3-8B           & 66.32 & -11.40 & \textbf{62.35} & 22.54 \\
        Llama3-70B          & 55.44 & -25.20 & 60.44 & 31.73 \\
        LLMCompiler-7B      & 68.82 & -9.48 & 53.53 & 14.48 \\
        LLMCompiler-13B     & 65.88 & -9.32 & 60.74 & 19.71 \\
        CodeBERT            & \textbf{70.88} & \textbf{-3.22} & 61.91 & \textbf{41.28} \\
        \bottomrule
    \end{tabular}
\end{table}

\section{Algorithm Classification}
\label{section:ac}
This section evaluates the proposed models on algorithm classification, a representative
program analysis task, comparing GNNs, LLMs, and state-of-the-art approaches.

\subsection{Dataset}

We use the POJ-104 dataset~\cite{lili-mou-2016} for a classification task, where a given
program is classified as a solution to one of 104 programming problems. Each problem has
500 corresponding C/C++ solutions, for a total of 52,000 programs. To generate LLVM IR 
code from the programs, we add include directives, replace \texttt{void main} with 
\texttt{int main}, and compile using \texttt{clang14 -O3} as C, retrying as C++ if 
needed.

\subsection{GNN Preprocessing}

CFGs are built from the LLVM IR with instruction count and autophase features as in 
Section~\ref{section:hdm_gnn_pre}, but no graph-level features are included.

\subsection{LLM Preprocessing}

The original source code is used without preprocessing. The first five lines of the LLVM
IR are removed along with string literals containing the filename to prevent label 
leakage. Embeddings are generated as in Section~\ref{section:hdm_llm_pre} but no global 
features are added.

\subsection{Train-Validation-Test Split}

We use the train-validation-test split introduced in~\cite{tal-ben-nun-2018} for 
consistency with prior work. After excluding programs that do not compile, there are 
26,911 training, 8,921 validation, and 9,008 test examples.

Each hyperparameter combination is trained using early stopping (patience 10 epochs, 
minimum delta 0.0001) for up to 250 epochs on the training set. The model with maximum 
validation MCC is retrained for a fixed number of epochs on the combined training and 
validation sets. The number of training epochs is scaled according to the ratio of the 
combined dataset size (training + validation) to the original training set size. The 
model is evaluated on the test set, and accuracy is reported. The training procedure is 
otherwise consistent with Section~\ref{section:hdm_str_ncv}.

\paragraph{GNN Hyperparameters}

We evaluate the same GNN models as in Section~\ref{section:hdm_gnn_hps} using an 
identical hyperparameter grid, except for the dropout rate (0.5) and GNN hidden size
(128/512).

\paragraph{LLM Hyperparameters}

We evaluate the same LLM models as in Section~\ref{section:hdm_llm_hps} using an 
identical hyperparameter grid, except for the LSTM hidden size (128/512).

\paragraph{Results}

LLM-based models consistently outperform GNN-based models in algorithm classification, 
as shown in Table~\ref{tab:ac-error}. Unlike in heterogeneous device mapping, where
decoder-only LLMs performed similarly, LLMCompiler surpasses the other LLMs in algorithm
classification.

One possible explanation is that LLMCompiler's pretraining on a compiler emulation task,
which involves translating one LLVM IR program into an equivalent LLVM IR program, helps
it generalize better to algorithm classification. Since this task requires grouping 
programs that compute the same output, LLMCompiler's ability to learn structural 
similarities between functionally equivalent programs makes it more adept at recognizing
algorithmic equivalence. 

In contrast, the LLMCompiler models do not outperform on heterogeneous device mapping, 
where programs within a class are not necessarily functionally equivalent. In this 
context, functionally equivalent programs might be categorized into different classes 
based on performance differences, such as one being faster on the CPU while another 
excels on the GPU. This observation further suggests that LLMCompiler's pretraining is 
particularly effective for algorithm classification, where recognizing functional 
equivalence is crucial. It also implies that an LLM trained with specialized code 
optimization objectives could generate more effective embeddings.

\begin{table}[htbp]
    \centering
    \caption{Performance on algorithm classification}
    \label{tab:ac-error}
    \begin{tabular}{ccc}
        \toprule
        Model               & Error Rate \\
        \midrule
        PNA                 & 4.05 \\
        GAT                 & 6.48 \\
        \midrule
        CodeLlama-7B        & 2.11 \\
        CodeLlama-70B       & 2.31 \\
        Llama3-8B           & 2.21 \\
        Llama3-70B          & 2.53 \\
        LLMCompiler-7B      & 1.62 \\
        LLMCompiler-13B     & \textbf{1.54} \\
        CodeBERT            & 5.63 \\
        \bottomrule
    \end{tabular}
\end{table}

\subsection{State-of-the-Art Approaches}

To compare to the state-of-the-art, we select the LLM and GNN models that achieved the 
best test error rates. The top-performing LLM model, LLMCompiler-13B, achieves a new 
state-of-the-art error rate of 1.54\%, as shown in Table~\ref{tab:ac-others}.

\begin{table}[htbp]
    \centering
    \caption{Comparison with other methods}
    \label{tab:ac-others}
    \begin{tabular}{ccc}
        \toprule
        Model                                       & Error Rate \\
        \midrule
        GNN (ours)                                  & 4.05 \\
        LLM (ours)                                  & \textbf{1.54} \\
        \midrule
        inst2vec~\cite{tal-ben-nun-2018}            & 5.17 \\
        OSCAR~\cite{dinglan-peng-2021}              & 1.92 \\
        ProGraML~\cite{chris-cummins-2021}          & 3.33 \\
        FAIR~\cite{changan-niu-2024}                & 1.75 \\
        PerfoGraph~\cite{jamsaz-ali-2024}           & 5.00 \\ 
        \bottomrule
    \end{tabular}
\end{table}

\subsection{Ablation}

Removing the source code degrades all models, especially those not trained on LLVM IR 
code, while LLMCompiler is less affected. Likewise, removing LLVM IR code harms most
models, with LLMCompiler impacted the most. Table~\ref{tab:ac_feat_abl} shows the error 
rates for this ablation experiment.

\begin{table}[htbp]
    \centering
    \caption{Input features ablation}
    \label{tab:ac_feat_abl}
    \begin{tabular}{ccccc}
        \toprule
        & \multicolumn{2}{c}{No Source Code} & \multicolumn{2}{c}{No LLVM IR} \\
        \cmidrule(r){2-3}
        \cmidrule(r){4-5}
        
        Model               & Error Rate & Change & Error Rate & Change \\
        \midrule
        CodeLlama-7B        & 4.27 & 102.37 & 3.38 & 60.19 \\
        CodeLlama-70B       & 4.37 & 89.18  & \textbf{2.00} & \textbf{-13.42}\\
        Llama3-8B           & 5.68 & 157.01 & 2.19 & -0.90 \\
        Llama3-70B          & 4.40 & 73.91 & 2.25 & -11.07 \\
        LLMCompiler-7B      & 2.47 & 52.47 & 2.92 & 80.25 \\
        LLMCompiler-13B     & \textbf{2.33} & \textbf{51.30} & 2.16 & 40.26 \\
        CodeBERT            & 9.80 & 74.38 & 8.18 & 45.55 \\
        \bottomrule
    \end{tabular}
\end{table}

These results suggest that incorporating both source and LLVM IR code is crucial for 
optimal performance. LLMCompiler-13B, trained on both, performs best, highlighting
their complementary roles: source code conveys high-level algorithmic intent, while LLVM
IR code captures low-level structural patterns. This underscores the benefit of training
LLMs on both code representations.

We also experimented with replacing LSTM aggregation with unweighted mean aggregation, 
which led to a significant increase in error rates across all models, as shown in 
Table~\ref{tab:ac_lstm_abl}. This suggests modeling long range dependencies across 
chunks is required for optimal results.

\begin{table}[htbp]
    \centering
    \caption{LSTM ablation}
    \label{tab:ac_lstm_abl}
    \begin{tabular}{ccc}
        \toprule
        Model               & Error Rate & Change \\
        \midrule
        CodeLlama-7B        & \textbf{2.86} & \textbf{35.55} \\
        CodeLlama-70B       & 5.54 & 139.83 \\
        Llama3-8B           & 3.10 & 40.27 \\
        Llama3-70B          & 3.71 & 46.64 \\
        LLMCompiler-7B      & 2.92 & 80.25 \\
        LLMCompiler-13B     & 5.98 & 288.31 \\
        CodeBERT            & 8.32 & 47.78 \\
        \bottomrule
    \end{tabular}
\end{table}

\section{Conclusion and Future Work}
\label{section:conclusion}

In this work, we proposed a simple yet effective approach for embedding both source code
and LLVM IR using pretrained LLMs. To address the constraints of fixed LLM context 
windows, we split programs into non-overlapping chunks, independently embedded each 
chunk, and aggregated the resulting embeddings into a single program embedding using an 
LSTM network. Despite its simplicity, our method achieved an error rate of 1.54\% in 
algorithm classification, or a 12\% improvement over the state-of-the-art, as well as 
competitive results in heterogeneous device mapping. Remarkably, this was accomplished
without engineered graph representations, highlighting the potential of 
LLMCompiler-based embeddings for code analysis and optimization. 

Our findings suggest several promising directions for future research. First, our 
ablation studies showed that combining source and LLVM IR code embeddings can yield 
better performance compared to using either embedding alone. Moreover, LLMCompiler 
performed exceptionally well on algorithm classification, an analysis task aligned with 
its source-to-source pretraining objectives, but did not outperform on heterogeneous 
device mapping, an optimization task. This suggests training or fine-tuning a 
specialized LLM with performance-aware objectives for embedding LLVM IR code. We plan to
train such an LLM on a large corpus of LLVM IR code in future work.

Second, the simplicity of our approach affords numerous opportunities for enhancing 
embedding quality. Future research could explore alternative methods for constructing 
chunk embeddings, such as those discussed in~\cite{chankyu-lee-2025}, or investigate 
replacing LSTM-based chunk aggregation with other context extending mechanisms like 
RoPE~\cite{jianlin-su-2023}. Additional preprocessing, such as cleaning the source and 
LLVM IR code, trying multiple compiler flags or even including explicit prompts, could 
yield further performance gains. In short, our method is poised to benefit from ongoing 
advancements in techniques for extracting embeddings from LLMs.

Finally, while our current approach is computationally expensive, its cost could be 
significantly reduced through methods like pruning, quantization, or knowledge 
distillation. These optimizations would make our method more practical for real-world 
applications. The cost-effectiveness of our approach will continue to scale with 
reductions in LLM inference costs.

\section*{Acknowledgment}

This work was partially funded by an unrestricted gift from Google. We also thank Edniel 
Campos for his work on data preprocessing and hyperparameter sweeps. 

ChatGPT\footnote{https://openai.com/chatgpt} was used for editing assistance throughout
all sections of this paper.

\printbibliography

\end{document}